\documentclass[letterpaper]{article} 
\usepackage{aaai23}  
\usepackage{times}  
\usepackage{helvet}  
\usepackage{courier}  
\usepackage[hyphens]{url}  
\usepackage{graphicx} 
\usepackage{natbib}  
\usepackage{caption} 
\usepackage[utf8]{inputenc}
\usepackage{algorithm}
\usepackage{algorithmic}
\usepackage{multirow}
\usepackage{booktabs}
\usepackage{newfloat}
\usepackage{listings}
\DeclareCaptionStyle{ruled}{labelfont=normalfont,labelsep=colon,strut=off} 
\floatstyle{ruled}
\newfloat{listing}{tb}{lst}{}
\floatname{listing}{Listing}
\usepackage[dvipsnames]{xcolor}
\definecolor{darkblue}{rgb}{0, 0, 0.5}
\usepackage[colorlinks=true,citecolor=darkblue,linkcolor=darkblue,urlcolor=darkblue]{hyperref}

\title{Multi-Timescale Modeling of Human Behavior}
\author{
    Chinmai Basavaraj,\textsuperscript{\rm 1}
    Adarsh Pyarelal,\textsuperscript{\rm 2}
    Evan C Carter\textsuperscript{\rm 1}
}
\affiliations{
    \textsuperscript{\rm 1}Human Research and Engineering Directorate, United States Army Research Laboratory\\
    \textsuperscript{\rm 2}School of Information, University of Arizona\\

}

\usepackage{bibentry}

\begin{document}

\maketitle

\begin{abstract}
In recent years, the role of artificially intelligent (AI) agents has evolved from being basic tools to socially intelligent agents working alongside humans towards common goals.
In such scenarios, the ability to predict future behavior by observing past actions of their human teammates is highly desirable in an AI agent.
Goal-oriented human behavior is complex, hierarchical, and unfolds across multiple timescales.
Despite this observation, relatively little attention has been paid towards using multi-timescale features to model such behavior.
In this paper, we propose an LSTM network architecture that processes behavioral information at multiple timescales to predict future behavior.
We demonstrate that our approach for modeling behavior in multiple timescales substantially improves prediction of future behavior compared to methods that do not model behavior at multiple timescales.
We evaluate our architecture on data collected in an urban search and rescue scenario simulated in a virtual Minecraft-based testbed, and compare its performance to that of a number of valid baselines as well as other methods that do not process inputs at multiple timescales.
\end{abstract}

\section{Introduction}

Artificial Intelligence (AI) has become an integral part of people's daily lives.
AI has played a key role in driving innovations in fields such as healthcare \cite{toosizadeh_screening_2019}, banking \cite{netzer_when_2019}, military applications \cite{ch_artificial_1996}, and space exploration \cite{hedberg_ai_1997}.
People are accustomed to relying on AI as tools to aid them in their daily activities ranging from scheduling to driving.
In recent years, the role of AI has progressed from being tools to socially intelligent agents.
Human-Machine Teaming (HMT) is a popular area of research where AI agents are designed to work alongside human teammates to achieve common goals.

To be an effective teammate, an AI agent needs to be efficient at understanding, identifying, and predicting human behavior.
AI agents capable of accurately predicting future behavior by observing the past can intervene and direct a team to be more efficient and enhance team performance.
Such AI is highly sought after and have many real-world applications in areas such as game design, biomedical engineering \cite{cui_multi-scale_2016}, and autonomous driving \cite{gopinath_hmiway-env_2022}.

Human behavior is a complex process.
In goal oriented tasks such as USAR, human behaviour has a hierarchical structure in association with short-term and long-term goals which unfold across multiple timescales.
In order to accurately predict the future behaviour in such tasks, the model must be able to understand the hierarchical structure of human behavior.
However, relatively little attention has been paid to modeling human behavior at multiple timescales.
Models that incorporate features evaluated over multiple timescales are shown to perform better than models that tend to ignore them
in applications such as driver drowsiness detection through facial videos \cite{massoz_multi-timescale_2018}, automated speech recognition from raw speech signals \cite{takeda_multi-timescale_2018}, and text classification \cite{liu_multi-timescale_2015}.

We present an LSTM network architecture that processes human behavioural information at multiple timescales to predict future behavior from past observations.
We take inspiration from the works of \citet{hihi_hierarchical_1995}, \citet{koutnik_clockwork_2014} and \citet{liu_multi-timescale_2015} who designed LSTMs  with delayed connections and units operating at different timescales.
Our LSTM model takes two minutes of behavioral data as input and predicts thirty seconds of future behavior.
Our results show that the LSTM model processing behavioral data at multiple timescales performs substantially better at predicting future behavior compared to the LSTM model that does not utilize multi-timescale modeling.
We also compare our LSTM model performances to valid baseline measures that account for biases in the behavioral data such as class imbalance and biases resulting from the structure and design of the experiment.
We test our hypothesis in an urban search and rescue (USAR) scenario simulated in a virtual Minecraft-based testbed that is designed as a part of DARPA’s ASIST program \citep{huang_freeman_cooke_colonna-romano_wood_buchanan_caufman_2022}.
In this scenario, the environment is dynamic with in-game perturbations, and the possibility of civilians dying if they are not rescued promptly.
These elements make USAR a very demanding task and at the same time, a realistic and important testbed for adaptive AI development \cite{traichioiu_hierarchical_2015}.

The paper is organized as follows.
First, we discuss related work in the area followed by a description of the Minecraft USAR mission data. We also describe the process for defining player behavior in Minecraft.
Next, we discuss the evaluation of baseline measures and the detailed architecture of the multi-timescale LSTM model.
Finally, we provide the results and conclude with a discussion.

\section{Related Work}

\subsection{Human Behavior in Minecraft}

The DARPA's ASIST program developed a Minecraft based testbed that simulates a virtual USAR scenario \cite{huang_freeman_cooke_colonna-romano_wood_buchanan_caufman_2022} to develop AI teammates that have the capability to infer mental states, predict future needs, and intervene effectively in a socially-aware manner in order to perform as adaptable and resilient AI teammates.
Research has shown that having AI as teammates makes a significant impact in tasks such as USAR by effectively improving optimal path planning \cite{ch_artificial_1996}, response times \cite{mehmood_multi_2018}, and contextual assessment \cite{merino_cooperative_2005}.
Interactive virtual worlds such as Minecraft have shown huge potential for the study and analysis of human behavior \citep{paulo_2021}.
Multiple frameworks have been developed and are readily available to facilitate systematic collection and analysis of Minecraft in-play behavioral data \cite{muller_statistical_2015}.

\citet{muller_statistical_2015} used data from a Minecraft-based testbed to classify high-level player behavior relevant to Minecraft such as building, exploring, fighting, and mining, from a series of low-level game play observations.
Their classifier used data accumulated within a two-minute sliding window as input.
The two-minute window was selected based on the assumption that Minecraft specific tasks took approximately two minutes to complete.
Building on their work, \citet{saadat_explaining_2020} defined behavior in Minecraft as a temporal sequence of low-level events and then classified Minecraft event sequences to infer high-level player behavior.
Their approach relied more on the order of events rather than their frequency of occurrence by replacing consecutive repetitive events by a single event.

\subsection{Future Prediction}
Predicting the future is an extensively studied topic in the field of machine learning.
It is critical in many real-world applications to be able to accurately predict the future based on past observations.
A recent review by \citet{zhou_deep_2020} suggest the convolutional LSTM (ConvLSTM) model to be highly suitable for future prediction in many real-time applications.
\citet{shi_convolutional_2015} proposed a convolutional LSTM model for weather forecasting that outperformed state-of-the-art algorithms.
Research work by \citet{saxena_clockwork_2021} proposed Clockwork Variational Autoencoder (VAE) architecture that processed input video frames at multiple timescales and outperformed other top models in long-term video prediction.
\citet{liang_2021} extracted spatio-temporal features at different resolutions to predict player behavior in a Minecraft simulated USAR scenario.

\subsection{Multi-Timescale Modeling}
Models that incorporate multi-timescale features have been shown to perform better than models that tend to ignore them in several scenarios.
Similar to \citet{liang_2021}, \citet{traichioiu_hierarchical_2015} developed a two-layered approach to modeling player behavior in a simulated USAR scenario \cite{kitano_robocup_1999}.
The two-layered approach considered macro-level behavior responsible for strategic high-level decisions and micro-level behavior dealing with the local particularities and rapid environmental changes in the mission.
\citet{hihi_hierarchical_1995} were the first to propose a hierarchical Recurrent Neural Network (RNN) architecture to better capture long-term dependencies in a sequence. Their model architecture included delays and processed sequential data at multiple timescales.
Following their work, \citet{koutnik_clockwork_2014} proposed a Clockwork RNN model, where the hidden layer was partitioned into separate modules, and each module processed inputs at multiple timescales. The Clockwork RNN model improved performance significantly in audio signal generation, spoken word classification, and online handwriting recognition.
\citet{takeda_multi-timescale_2018} used multi-timescale features for automated speech recognition from raw speech signal and showed that a multi-timescale approach reduced the word error rate.
Using multi-timescale features to analyze face videos of drivers showed improved performance in driver drowsiness detection \citep{massoz_multi-timescale_2018}.
\citet{liu_multi-timescale_2015} developed a multi-timescale LSTM (MT-LSTM) architecture for modeling long sentences in a document and it outperformed other neural network architectures that did not rely on using multi-timescale features in a text classification task.

\begin{figure}[t]
\centering
\includegraphics[width=0.95\columnwidth]{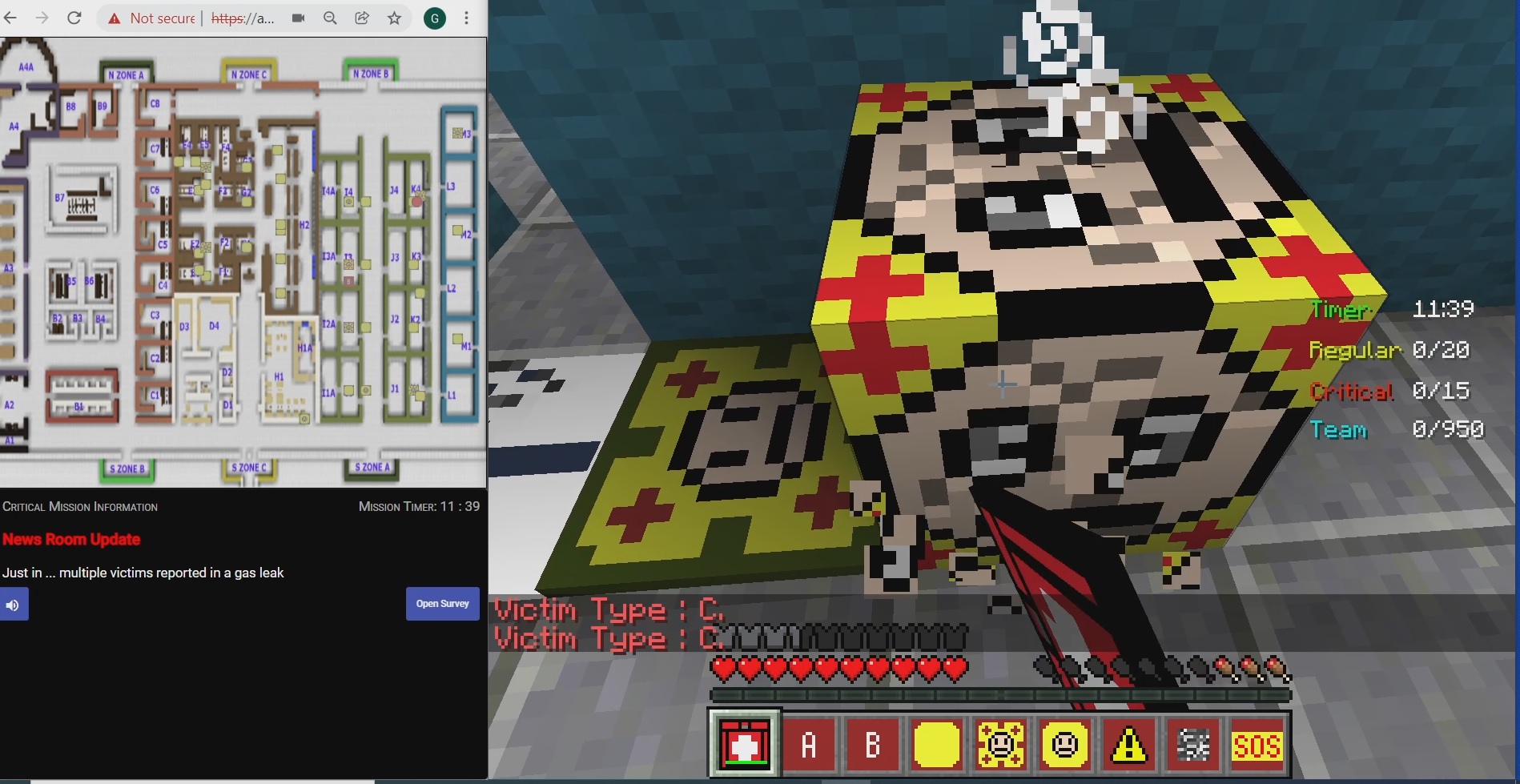}
\caption{A screenshot of the USAR Minecraft mission shows a medic field of view triaging a critical victim placed next to a marker. On the lower-right corner is the inventory. On the upper-left corner is the dynamic map. Below the map, players can see mission critical information.}
\label{fig1}
\end{figure}

\section{Experiment and Data}
In each experiment session, teams of three qualified participants were engaged in a simulated USAR mission in Minecraft \cite{huang_freeman_cooke_colonna-romano_wood_buchanan_caufman_2022}.
First, the participants received training that introduced them to the rules of the game and provided some hands-on experience with the Minecraft environment.
Next, the participants were engaged in two USAR missions with different map configurations.
The participants were required to coordinate and work together as a team to complete the mission.

\paragraph{Roles} In each mission, the participants were assigned a unique role: medic, transporter, or engineer.
Each role was equipped with a specific set of tools, which they had to select from their inventory in order to perform role specific actions \cite{pyarelal_2022}.
The medic could triage victims, the engineer could clear rubble to rescue trapped players and victims, and the transporter could use the signal to detect victims without entering the room.
All roles can transport triaged victims to treatment areas to be rescued.
Each role also had different attributes such as color sign and walking speed.

\paragraph{Victims}
There were three types of victims in the SAR mission - victims with abrasions (type A), bone-damage (type B), and critical  victims (type C).
There were 20 regular victims and 15 critical victims.
The victims had to be transported to their respective treatment area types to be rescued.
While type A and B victims were worth 10 points and could be rescued by a single player, type C victims were worth 50 points and required at least two players to rescue. This design element was implemented to encourage teamwork.

\paragraph{Communication} The participants could communicate with each other during the mission through audio and also had access to a dynamic map and mission critical information that was provided before the start of the mission.
This information was unique to each role and the participants were expected to communicate and share the information.
Each mission was preceded by a two-minute planning phase.
The goal of the mission was to gain as many points as possible by rescuing the victims within the time limit of 15 minutes.

\begin{table}
    \centering
    \begin{tabular}{lcc}
        \toprule
        Behavior (actions) & Class Label & Alphabet \\
        \midrule
        Stationary & 0 & ST \\
        Navigate (corridor) & 1 & NV \\
        Search (room) & 2 & SR \\
        Open door & 3 & OD \\
        Transport victim & 4 & TV \\
        Place marker & 5 & PM \\
        Remove marker & 6 & RM \\
        Tool used & 7 & TU \\
        Role specific action & 8 & RA \\
        Item equipped & 9 & IE \\
        Audio communication & 10 & AC\\
        \bottomrule
    \end{tabular}
    \caption{List of semantic labels that represent low-level Minecraft states and actions.}
    \label{table:1}
\end{table}

We analyzed data from 115 teams collected as a part of the ASIST Study 3 \cite{huang_freeman_cooke_colonna-romano_wood_buchanan_caufman_2022}.
Each team had three participants and conducted two missions each.
We further segmented each player's behavioral data associated with a single mission into two-minute sliding windows\footnote{We chose the window length of two minutes based on the average duration required to complete a majority of the tasks in the mission \cite{muller_statistical_2015}.}
with a step size of 30-seconds, obtaining approximately 21000 data points in total, where each data point corresponds to a player's behavior for two minutes in a specific mission.

\subsection{Data Representation}

We define behavior in the mission as a sequence of events associated with each participant during the mission.
To effectively capture the participants' state in the mission, we defined the set of features listed in Table \ref{table:2}.
The Minecraft testbed was instrumented to send messages that provided information about every player's actions and states in the virtual environment every 100 milliseconds.
We translated the received messages to compute the values for all the listed features.
The participants' behavior in the mission can be viewed as a multivariate time-series of evaluated features $T= \{T_1, T_2,T_3, ... , T_n \}$ where $T_1$ represents a vector of feature values evaluated at time $t_1$.

By thoroughly observing every participant's behavior during the missions, we also defined a set of semantic labels that represent low-level actions and states, in order to make behavior representations more human-readable and easier to interpret.
Additionally, prior research suggests that, making predictions at a semantic level is more effective and beneficial for AI teammate compared to predicting raw feature values \cite{cui_multi-scale_2016}.
Table \ref{table:1} lists the set of semantic labels defined.
We assigned a semantic label for every feature vector evaluated at a specific time.
Thus, we can represent the player behavior in a mission as a sequence of semantic labels.
For example, a player's behavior could be represented as $S$ = \{NV, NV, OD, SR, SR, ..., TV, NV \}.

\begin{table}
    \centering
    \begin{tabular}{l l}
        \toprule
        Feature & Value type \\
        \midrule
        CurrentLocation & Categorical \\
        CurrentVelocity & Numerical \\
        CriticalVictimsTriaged & Numerical \\
        CriticalVictimsSaved & Numerical \\
        DistanceTraveled & Numerical \\
        ItemEquipped & Categorical \\
        MissionTime & Numerical \\
        PlayerRole & Categorical \\
        ProximityToNearestDoor & Numerical \\
        ProximityToNearestTreatmentArea & Numerical \\
        ProximityToMedic & Numerical\\
        ProximityToEngineer & Numerical\\
        ProximityToTransporter & Numerical\\
        ProximityToNearestRegular & Numerical\\
        ProximityToNearestCritical & Numerical\\
        ProximityToNearestMarker & Numerical \\
        RegularVictimsTriaged & Numerical \\
        RegularVictimsSaved & Numerical \\
        ToolUsed & Categorical \\

        \bottomrule
    \end{tabular}
    \caption{Features evaluated to represent player's state in the Minecraft USAR mission. The features were evaluated at multiple timescales.}
    \label{table:2}
\end{table}

\section{Approach}

\begin{figure}[t]
\centering
\includegraphics[width=0.9\columnwidth]{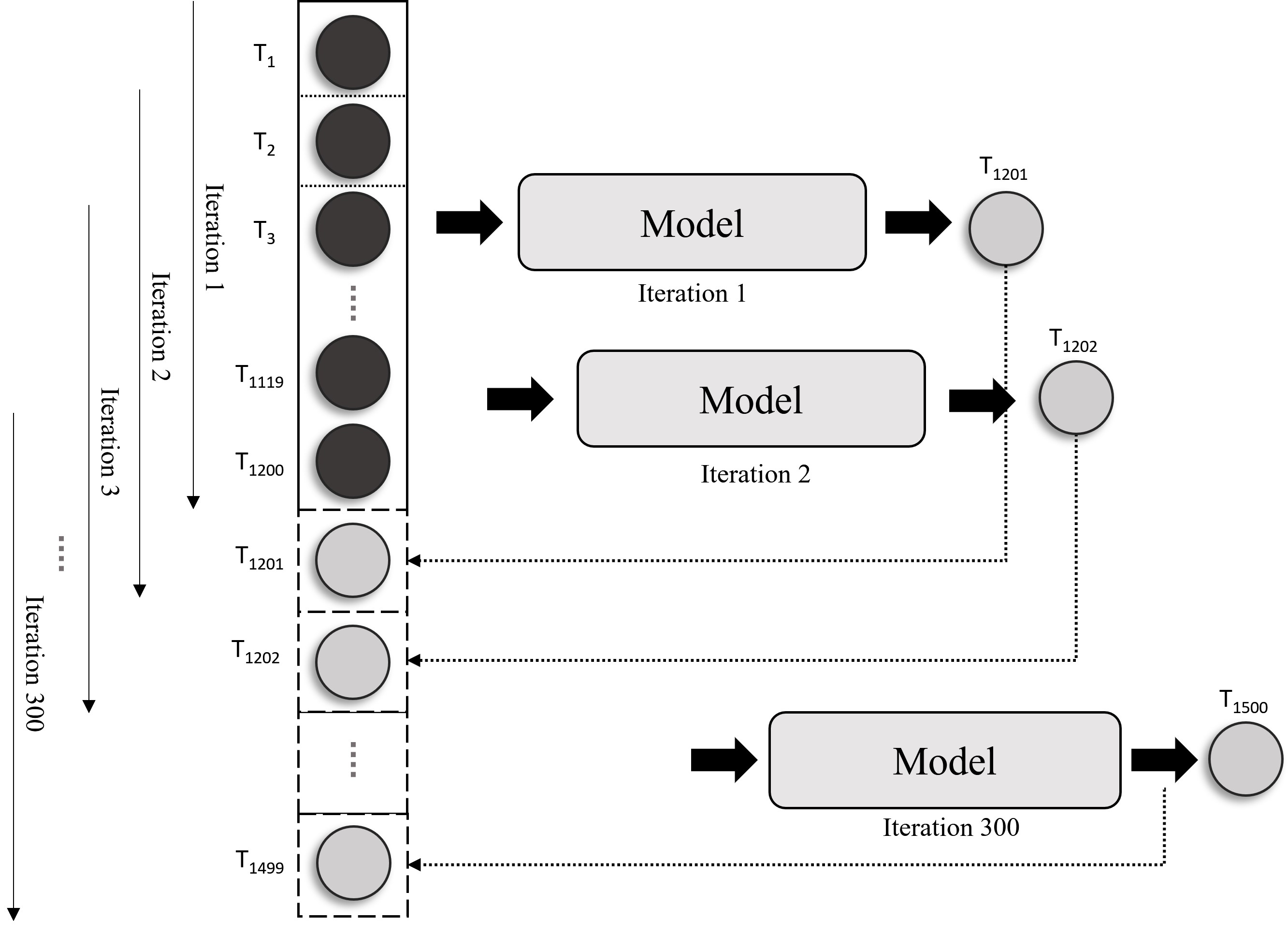}
\caption{Iterative process of predicting future behavior from a sequence of past behavioral data}
\label{fig2}
\end{figure}

Our goal is to predict future behavior from past observations.
Our models take a multivariate time-series of feature vectors (Table \ref{table:2}) associated with two minutes of behavioral data $T= \{T_1,  T_2, T_3, ... , T_{n} \}$ as input and predict a feature vector associated with the next time step $T_{n+1}$ step as the output.
The predicted feature vector from the first iteration $T_{n+1}$ is squeezed into the sequence of input feature vectors and provided as the input for the next iteration.
Figure \ref{fig2} depicts the process of future behavior prediction.
We repeat the iterative process for 300 iterations, which correspond to 30 seconds of future behavior, as our data points are sampled at a rate of 100 milliseconds.

After the final iteration, we map the sequence of output feature vectors to a sequence of semantic labels defined in Table \ref{table:1}.
We separately train a densely connected sequence-to-sequence LSTM model to do the mapping.
We assume that the 2-minute behavioral data points obtained from the same participant and the same mission are independent.
We evaluate the performance of the model on predicting future behavior by measuring the prediction accuracy of the generated sequence of semantic labels against the ground-truth.
We employ 10-fold cross validation to evaluate the performance and report the mean accuracy and standard error.

\subsection{Model}
In this section we describe the architecture of the multi-timescale LSTM model.
The model architecture is shown in Figure 3.
LSTM models were originally proposed by \citet{hochreiter_long_1997}, and are considered to be the state-of-the-art for modeling time-series data.
Inspired by the works of \citet{hihi_hierarchical_1995}, \citet{koutnik_clockwork_2014}, and \citet{liu_multi-timescale_2015}, we slightly modified the LSTM architecture to have the hidden layer units update at different time intervals, thus effectively having them operate at different timescales.

The hidden nodes of the LSTM are grouped into $k$ groups $\{G_1,G_2, ... , G_k\}$.
Each group $G_i,(1\leq i \leq k)$ of hidden nodes is activated at different time periods $t_i$.
In our case, we group the hidden nodes into 3 groups.
The gates and the weight matrices of the LSTM network are also partitioned accordingly to the corresponding hidden nodes group.
One group of hidden nodes along with the weight matrices operates the same way as a standard LSTM network.
At each time step $t$, only the groups $G_i$ that satisfy the condition $(t\   mod\   T_i) = 0$ are executed.
Similarly, when we train the network using back-propagation, the error propagation follows the same rule. The error of non-activated group of hidden nodes gets copied back in time

We train the multi-timescale LSTM model to minimize the root mean squared error (RMSE) of the predicted values of features to the true  values.
The number of LSTM units in the hidden layer node is set to 64.
We set the base learning rate to be 0.001 and train our network for 50 epochs.
We set the batch size as 64 and use Adam as the optimization algorithm.

The output of the multi-timescale LSTM model is a sequence of feature vectors corresponding to 30 seconds of predicted future behavior.
In order to map the sequence of feature vectors to a sequence of semantic labels, we train another LSTM network separately.
For the sequence-to-sequence mapping LSTM model, we train the model to reduce the categorical cross-entropy loss between predicted labels to the ground truth labels.
We set the batch size as 32 and train the model for 20 epochs.
The number of LSTM units in the hidden layer node is set to 64.
We set the base learning rate to be 0.001 and train the model using the Adam optimization algorithm.

\begin{figure}[t]
\centering
\includegraphics[width=0.9\columnwidth]{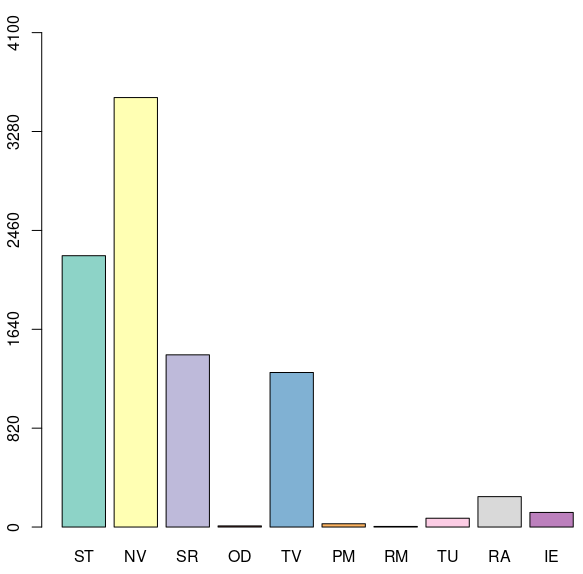}
\caption{Histogram of distribution of semantic labels from data  of the multi-timescale LSTM model.}
\label{fig3}
\end{figure}

\begin{figure*}[t]
\centering
\includegraphics[width=1.9\columnwidth]{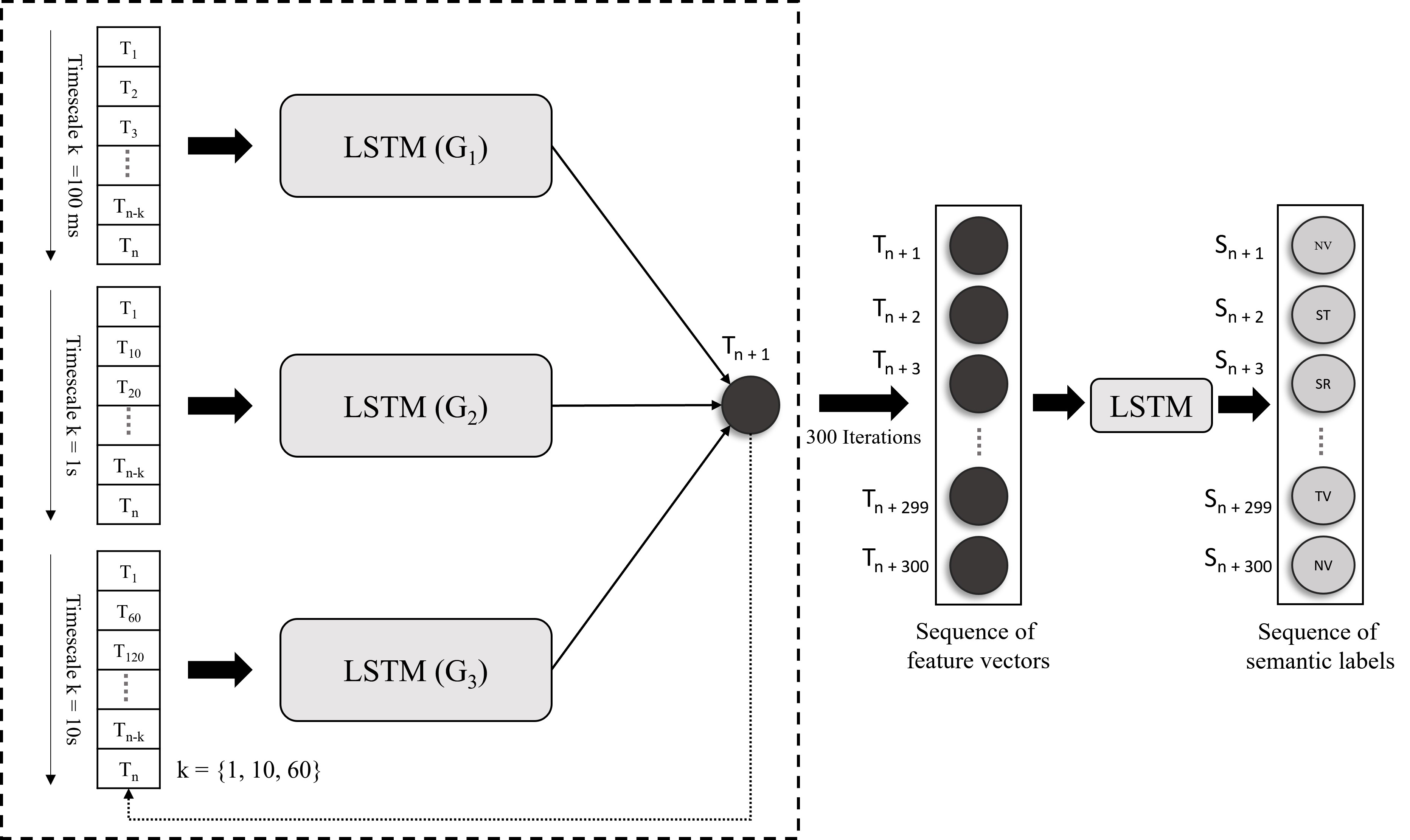}
\caption{Architecture of the multi-timescale LSTM model. The hidden layer units of the LSTM network are grouped into 3 groups that update at 3 different timescales: 100ms, 1s, and 10s. After the final iteration, the model outputs a sequence of feature vectors corresponding to 30 seconds of future behavior which are then mapped to a sequence of semantic labels.}
\label{fig4}
\end{figure*}

\section{Results}

\subsection{Baseline}
In this section we describe the baseline measure used to compare the performance of the multi-timescale LSTM model in predicting future behavior.
In the mission, the participants spent a majority of their time being stationary, searching the room, or navigating the hallways.
Figure \ref{fig3} shows the distribution of semantic labels in our dataset \cite{gabadinho_analyzing_2011}.
A na\"ive algorithm that predicts the most common semantic label in the dataset, which is navigating the corridors (NV) as output at every time step $t$, it will yield a 53\% prediction accuracy.
We call this measure baseline 1.
Given that we have 11 different class labels, the na\"ive chance percentage would be approximately 9\%.
We notice that the Baseline 1 measure is significantly higher than chance.

\subsection{Multi-Timescale Features}
We compare the performance of the multi-timescale LSTM model in predicting future behavior against an LSTM model that does not utilize multi-timescale features.

\begin{table}
    \centering
    \begin{tabular}{l c}
        \toprule
        Model & Prediction Accuracy (\%)\\
        \midrule
        Baseline 1 & 53.05 $\pm$ 2 \\
        LSTM &  66.60 $\pm$ 2\\
        Multi-Timescale LSTM & 71.24 $\pm$ 2  \\
        \bottomrule
    \end{tabular}
    \caption{Performance of models in the prediction of future behavior by observing the past. }
    \label{table:3}
\end{table}

We notice that both the LSTM and multi-timescale LSTM models perform better than baseline 1.
This might imply that the LSTM models are learning meaningful representations of human behavior from the sequence of defined features, and perform better at predicting future behavior compared to the na\"{i}ve algorithm employed in baseline 1.
The multi-timescale LSTM model shows about 7.8\% improvement in prediction of future behavior over the LSTM model that does not utilize multi-timescale features.
As hypothesized, incorporating multi-timescale features by updating the hidden layer units of the LSTM at different timescales results in better modeling and prediction of behavior than a straightforward LSTM network.
We conduct further analysis to find that increasing and decreasing the interval between timescales does not have a major impact on future  prediction accuracy. We obtain the best result by using timescale values of 100ms, 1s, and 10s.

\begin{table}
    \small
    \centering
    \begin{tabular}{llp{0.7in}}
        \toprule
        Role & Model & Prediction Accuracy (\%) \\
        \midrule
        Medic & Baseline 1 & 45.76 $\pm$ 3 \\
         & LSTM &  67.58 $\pm$ 2\\
         & Multi-Timescale LSTM & 74.08 $\pm$ 1  \\
        \midrule
        Engineer & Baseline 1 & 54.87 $\pm$ 2 \\
         & LSTM &  66.55 $\pm$ 2\\
         & Multi-Timescale LSTM & 70.20 $\pm$ 2  \\
        \midrule
        Transporter & Baseline 1 & 55.37 $\pm$ 2 \\
         & LSTM &  66.12 $\pm$ 2\\
         & Multi-Timescale LSTM & 70.58 $\pm$ 2  \\
        \bottomrule
    \end{tabular}
    \caption{Role-specific analysis of performance of models in predicting future behavior by observing the past.}
    \label{table:4}
\end{table}

\subsection{Role-Specific Analysis}
We notice clear differences in behavior associated with each role in the USAR mission - medic, engineer, and transporter.
The semantic label distribution, the frequency of label occurrence over time, and the transition between labels are distinct for each role.
Based on their assigned role, the players employ different strategies as the mission progresses.
We compute and compare the behavior prediction accuracy using only data from a specific role across baseline 1 measure, LSTM, and multi-timescale LSTM models.
Table \ref{table:4} shows the comparison of prediction accuracies for different roles.

We notice that, for a medic, the use of multi-timescale features boosts the accuracy of predicting future behavior (by 9.6\%) compared to that of an engineer (5.5\%) or that of a transporter (6.7\%).

\section{Conclusion}
We are able to predict the behavior of a participant engaged in a USAR mission 30 seconds into the future by observing two minutes of their past behavior,with an accuracy of 74 $\pm$ 1\%.
Having a group of hidden nodes in the LSTM update at different timescales allows the model to be more efficient at modeling and predicting human behavior in goal-oriented tasks such as USAR.
Incorporating multi-timescale features allows the model to effectively capture both short-term rapid changes in the environment as well as long-term strategic plans employed by the players.
Our results show that using multi-timescale features to model and predict human behavior in Minecraft simulated USAR task consistently outperforms the LSTM model that does not use multi-timescale features.
From role-specific analysis, we infer that incorporating multi-timescale features improves the behavior prediction for the medic roles better than the engineer and transporter.

\section{Acknowledgments}

Research was sponsored by the Army Research Office and was accomplished under
Grant Number W911NF-20-1-0002. The views and conclusions contained in this
document are those of the authors and should not be interpreted as representing
the official policies, either expressed or implied, of the Army Research Office
or the U.S. Government. The U.S. Government is authorized to reproduce and
distribute reprints for Government purposes notwithstanding any copyright
notation herein.

\bibliography{aaai23}

\begin{thebibliography}{26}
\providecommand{\natexlab}[1]{#1}

\bibitem[{Blitch(1996)}]{ch_artificial_1996}
Blitch, J.~G. 1996.
\newblock Artificial intelligence technologies for robot assisted urban search
  and rescue.
\newblock \emph{Expert Systems with Applications}, 11(2): 109--124.

\bibitem[{Cui, Chen, and Chen(2016)}]{cui_multi-scale_2016}
Cui, Z.; Chen, W.; and Chen, Y. 2016.
\newblock Multi-{Scale} {Convolutional} {Neural} {Networks} for {Time} {Series}
  {Classification}.
\newblock \emph{arXiv:1603.06995 [cs]}.
\newblock ArXiv: 1603.06995.

\bibitem[{Gabadinho et~al.(2011)Gabadinho, Ritschard, Müller, and
  Studer}]{gabadinho_analyzing_2011}
Gabadinho, A.; Ritschard, G.; Müller, N.~S.; and Studer, M. 2011.
\newblock Analyzing and {Visualizing} {State} {Sequences} in \textit{{R}} with
  {{TraMineR}}.
\newblock \emph{Journal of Statistical Software}, 40(4).

\bibitem[{Gopinath et~al.(2022)Gopinath, DeCastro, Rosman, Sumner, Morgan,
  Hakimi, and Stent}]{gopinath_hmiway-env_2022}
Gopinath, D.; DeCastro, J.; Rosman, G.; Sumner, E.; Morgan, A.; Hakimi, S.; and
  Stent, S. 2022.
\newblock {HMIway}-{Env}: {A} {Framework} for {Simulating} {Behaviors} and
  {Preferences} {To} {Support} {Human}-{AI} {Teaming} in {Driving}.
\newblock 4342--4350.

\bibitem[{Hedberg(1997)}]{hedberg_ai_1997}
Hedberg, S. 1997.
\newblock {AI} coming of age: {NASA} uses {AI} for autonomous space
  exploration.
\newblock \emph{IEEE Expert}, 12(3): 13--15.
\newblock Conference Name: IEEE Expert.

\bibitem[{Hihi and Bengio(1995)}]{hihi_hierarchical_1995}
Hihi, S.; and Bengio, Y. 1995.
\newblock Hierarchical {Recurrent} {Neural} {Networks} for {Long}-{Term}
  {Dependencies}.
\newblock In \emph{Advances in {Neural} {Information} {Processing} {Systems}},
  volume~8. MIT Press.

\bibitem[{Hochreiter and Schmidhuber(1997)}]{hochreiter_long_1997}
Hochreiter, S.; and Schmidhuber, J. 1997.
\newblock Long {Short}-{Term} {Memory}.
\newblock \emph{Neural Computation}, 9(8): 1735--1780.
\newblock Conference Name: Neural Computation.

\bibitem[{Huang et~al.(2022)Huang, Freeman, Cooke, Colonna-Romano, Wood,
  Buchanan, and
  Caufman}]{huang_freeman_cooke_colonna-romano_wood_buchanan_caufman_2022}
Huang, L.; Freeman, J.; Cooke, N.; Colonna-Romano, J.; Wood, M.~D.; Buchanan,
  V.; and Caufman, S.~J. 2022.
\newblock Exercises for Artificial Social Intelligence in Minecraft Search and
  Rescue for Teams.

\bibitem[{Kitano et~al.(1999)Kitano, Tadokoro, Noda, Matsubara, Takahashi,
  Shinjou, and Shimada}]{kitano_robocup_1999}
Kitano, H.; Tadokoro, S.; Noda, I.; Matsubara, H.; Takahashi, T.; Shinjou, A.;
  and Shimada, S. 1999.
\newblock {RoboCup} {Rescue}: search and rescue in large-scale disasters as a
  domain for autonomous agents research.
\newblock In \emph{{IEEE} {SMC}'99 {Conference} {Proceedings}. 1999 {IEEE}
  {International} {Conference} on {Systems}, {Man}, and {Cybernetics} ({Cat}.
  {No}.{99CH37028})}, volume~6, 739--743 vol.6.
\newblock ISSN: 1062-922X.

\bibitem[{Koutnik et~al.(2014)Koutnik, Greff, Gomez, and
  Schmidhuber}]{koutnik_clockwork_2014}
Koutnik, J.; Greff, K.; Gomez, F.; and Schmidhuber, J. 2014.
\newblock A {Clockwork} {RNN}.
\newblock In \emph{Proceedings of the 31st {International} {Conference} on
  {Machine} {Learning}}, 1863--1871. PMLR.
\newblock ISSN: 1938-7228.

\bibitem[{Liu et~al.(2015)Liu, Qiu, Chen, Wu, and
  Huang}]{liu_multi-timescale_2015}
Liu, P.; Qiu, X.; Chen, X.; Wu, S.; and Huang, X. 2015.
\newblock Multi-{Timescale} {Long} {Short}-{Term} {Memory} {Neural} {Network}
  for {Modelling} {Sentences} and {Documents}.
\newblock In \emph{Proceedings of the 2015 {Conference} on {Empirical}
  {Methods} in {Natural} {Language} {Processing}}, 2326--2335. Lisbon,
  Portugal: Association for Computational Linguistics.

\bibitem[{Massoz, Verly, and
  Van~Droogenbroeck(2018)}]{massoz_multi-timescale_2018}
Massoz, Q.; Verly, J.~G.; and Van~Droogenbroeck, M. 2018.
\newblock Multi-{Timescale} {Drowsiness} {Characterization} {Based} on a
  {Video} of a {Driver}’s {Face}.
\newblock \emph{Sensors}, 18(9): 2801.
\newblock Number: 9 Publisher: Multidisciplinary Digital Publishing Institute.

\bibitem[{Mehmood et~al.(2018)Mehmood, Ahmed, Kristensen, and
  Ahsan}]{mehmood_multi_2018}
Mehmood, S.; Ahmed, S.; Kristensen, A.~S.; and Ahsan, D. 2018.
\newblock Multi {Criteria} {Decision} {Analysis} ({MCDA}) of {Unmanned}
  {Aerial} {Vehicles} ({UAVs}) as a {Part} of {Standard} {Response} to
  {Emergencies}.
\newblock In \emph{Proceeding of the {International} {Conference} on {Green}
  {Computing} and {Engineering} {Technologies} 2018}. Gyancity International
  Publishers.

\bibitem[{Merino et~al.(2005)Merino, Caballero, Martinez-de Dios, and
  Ollero}]{merino_cooperative_2005}
Merino, L.; Caballero, F.; Martinez-de Dios, J.; and Ollero, A. 2005.
\newblock Cooperative {Fire} {Detection} using {Unmanned} {Aerial} {Vehicles}.
\newblock In \emph{Proceedings of the 2005 {IEEE} {International} {Conference}
  on {Robotics} and {Automation}}, 1884--1889.
\newblock ISSN: 1050-4729.

\bibitem[{Müller et~al.(2015)Müller, Kapadia, Frey, Klinger, Mann,
  Solenthaler, Sumner, and Gross}]{muller_statistical_2015}
Müller, S.; Kapadia, M.; Frey, S.; Klinger, S.; Mann, R.~P.; Solenthaler, B.;
  Sumner, R.~W.; and Gross, M. 2015.
\newblock Statistical {Analysis} of {Player} {Behavior} in {Minecraft}.
\newblock Conference Name: Foundations of Digital Games (FDG 2015) ISBN:
  9780991398249 Meeting Name: Foundations of Digital Games (FDG 2015) Place:
  Pacific Grove, California Publisher: Society for the Advancement of the
  Science of Digital Games.

\bibitem[{Netzer, Lemaire, and Herzenstein(2019)}]{netzer_when_2019}
Netzer, O.; Lemaire, A.; and Herzenstein, M. 2019.
\newblock When {Words} {Sweat}: {Identifying} {Signals} for {Loan} {Default} in
  the {Text} of {Loan} {Applications}.
\newblock \emph{Journal of Marketing Research}, 56(6): 960--980.
\newblock Publisher: SAGE Publications Inc.

\bibitem[{Pyarelal(2022)}]{pyarelal_2022}
Pyarelal, A. 2022.
\newblock ToMCAT (UAZ + TAMU) Study 3 Preregistration.

\bibitem[{Saadat and Sukthankar(2020)}]{saadat_explaining_2020}
Saadat, S.; and Sukthankar, G. 2020.
\newblock Explaining {Differences} in {Classes} of {Discrete} {Sequences}.
\newblock In \emph{2020 {IEEE}/{WIC}/{ACM} {International} {Joint} {Conference}
  on {Web} {Intelligence} and {Intelligent} {Agent} {Technology} ({WI}-{IAT})},
  129--136.

\bibitem[{Saxena, Ba, and Hafner(2021)}]{saxena_clockwork_2021}
Saxena, V.; Ba, J.; and Hafner, D. 2021.
\newblock Clockwork {Variational} {Autoencoders}.
\newblock In \emph{Advances in {Neural} {Information} {Processing} {Systems}},
  volume~34, 29246--29257. Curran Associates, Inc.

\bibitem[{Shi et~al.(2015)Shi, Chen, Wang, Yeung, Wong, and
  WOO}]{shi_convolutional_2015}
Shi, X.; Chen, Z.; Wang, H.; Yeung, D.-Y.; Wong, W.-k.; and WOO, W.-c. 2015.
\newblock Convolutional {LSTM} {Network}: {A} {Machine} {Learning} {Approach}
  for {Precipitation} {Nowcasting}.
\newblock In \emph{Advances in {Neural} {Information} {Processing} {Systems}},
  volume~28. Curran Associates, Inc.

\bibitem[{Soares, Pyarelal, and Barnard(2021)}]{paulo_2021}
Soares, P.; Pyarelal, A.; and Barnard, K. 2021.
\newblock Probabilistic Modeling of Human Teams to Infer False Beliefs.
\newblock In \emph{2021 AAAI Fall Symposium}.

\bibitem[{Takeda, Nakadai, and Komatani(2018)}]{takeda_multi-timescale_2018}
Takeda, R.; Nakadai, K.; and Komatani, K. 2018.
\newblock Multi-timescale {Feature}-extraction {Architecture} of {Deep}
  {Neural} {Networks} for {Acoustic} {Model} {Training} from {Raw} {Speech}
  {Signal}.
\newblock In \emph{2018 {IEEE}/{RSJ} {International} {Conference} on
  {Intelligent} {Robots} and {Systems} ({IROS})}, 2503--2510.
\newblock ISSN: 2153-0866.

\bibitem[{Toosizadeh et~al.(2019)Toosizadeh, Ehsani, Wendel, Zamrini, Connor,
  and Mohler}]{toosizadeh_screening_2019}
Toosizadeh, N.; Ehsani, H.; Wendel, C.; Zamrini, E.; Connor, K.~O.; and Mohler,
  J. 2019.
\newblock Screening older adults for amnestic mild cognitive impairment and
  early-stage {Alzheimer}’s disease using upper-extremity dual-tasking.
\newblock \emph{Scientific Reports}, 9(1): 10911.
\newblock Number: 1 Publisher: Nature Publishing Group.

\bibitem[{Trăichioiu and Visser(2015)}]{traichioiu_hierarchical_2015}
Trăichioiu, M.; and Visser, A. 2015.
\newblock Hierarchical {Decision} {Making} for {Search} and {Rescue}
  {Teamwork}.

\bibitem[{Zhang, Lieffers, and Pyarelal(2021)}]{liang_2021}
Zhang, L.; Lieffers, J.; and Pyarelal, A. 2021.
\newblock Using Features at Multiple Temporal and Spatial Resolutions to
  Predict Human Behavior in Real Time.
\newblock In \emph{2021 AAAI Fall Symposium}.

\bibitem[{Zhou, Dong, and El~Saddik(2020)}]{zhou_deep_2020}
Zhou, Y.; Dong, H.; and El~Saddik, A. 2020.
\newblock Deep {Learning} in {Next}-{Frame} {Prediction}: {A} {Benchmark}
  {Review}.
\newblock \emph{IEEE Access}, 8: 69273--69283.
\newblock Conference Name: IEEE Access.

\end{thebibliography}

\end{document}